\documentclass[sigconf]{acmart}
\usepackage{graphicx}
\usepackage{textcomp}
\usepackage{xcolor}

\usepackage{bm}
\usepackage{algorithmic} %算法表
\usepackage{algorithm}
\usepackage{tabularx}
\usepackage{subfigure}
\usepackage{booktabs} % For formal tables
\usepackage{multirow}

\AtBeginDocument{%
  \providecommand\BibTeX{{%
    \normalfont B\kern-0.5em{\scshape i\kern-0.25em b}\kern-0.8em\TeX}}}

\copyrightyear{2022}
\acmYear{2022}
\setcopyright{acmcopyright}
\acmConference[SIGIR '22] {Proceedings of the 45th International ACM SIGIR Conference on Research and Development in Information Retrieval}{July 11--15, 2022}{Madrid, Spain.}
\acmBooktitle{Proceedings of the 45th International ACM SIGIR Conference on Research and Development in Information Retrieval (SIGIR '22), July 11--15, 2022, Madrid, Spain}
\acmPrice{15.00}
\acmISBN{978-1-4503-8732-3/22/07}
\acmDOI{10.1145/3477495.3531733}  % https://doi.org/10.1145/3477495.3531733

\begin{document}
\fancyhead{}
\begin{sloppypar}  %处理text行溢出

\title{MetaCVR: Conversion Rate Prediction via Meta Learning in Small-Scale Recommendation Scenarios}

\author{Xiaofeng Pan, Ming Li}
\authornote{Both authors contributed equally to this paper.}
\email{pxfvintage@163.com, hongming.lm@alibaba-inc.com}
\affiliation{
  \institution{Alibaba Group}
  \country{China}
}

\author{Jing Zhang}
\email{jing.zhang1@sydney.edu.au}
\affiliation{
  \institution{The University of Sydney}
  \country{Australia}
}

\author{Keren Yu, Hong Wen, Luping Wang}
\email{{keren.ykr, qinggan.wh}@alibaba-inc.com, zjlxwlp8412@126.com}
\affiliation{
  \institution{Alibaba Group}
  \country{China}
}

\author{Chengjun Mao}
\author{Bo Cao}
\email{{chengjun.mcj, zhizhao.cb}@alibaba-inc.com}
\affiliation{
  \institution{Alibaba Group}
  \country{China}
}


\begin{abstract}
Different from large-scale platforms such as Taobao and Amazon, CVR modeling in small-scale recommendation scenarios is more challenging due to the severe \textbf{D}ata \textbf{D}istribution \textbf{F}luctuation (DDF) issue. DDF prevents existing CVR models from being effective since 1) several months of data are needed to train CVR models sufficiently in small scenarios, leading to considerable distribution discrepancy between training and online serving; and 2) e-commerce promotions have significant impacts on small scenarios, leading to distribution uncertainty of the upcoming time period. In this work, we propose a novel CVR method named MetaCVR from a perspective of meta learning to address the DDF issue. Firstly, a base CVR model which consists of a Feature Representation Network (FRN) and output layers is designed and trained sufficiently with samples across months. Then we treat time periods with different data distributions as different occasions and obtain positive and negative prototypes for each occasion using the corresponding samples and the pre-trained FRN. Subsequently, a Distance Metric Network (DMN) is devised to calculate the distance metrics between each sample and all prototypes to facilitate mitigating the distribution uncertainty. At last, we develop an Ensemble Prediction Network (EPN) which incorporates the output of FRN and DMN to make the final CVR prediction. In this stage, we freeze the FRN and train the DMN and EPN with samples from recent time period, therefore effectively easing the distribution discrepancy. To the best of our knowledge, this is the first study of CVR prediction targeting the DDF issue in small-scale recommendation scenarios. Experimental results on real-world datasets validate the superiority of our MetaCVR and online A/B test also shows our model achieves impressive gains of 11.92\% on PCVR and 8.64\% on GMV.
\end{abstract}




\maketitle

\section{Introduction}
As an essential part of recommender system, Conversion Rate (CVR) prediction \cite{wen2019multi} has been widely used in modern e-commerce and attracted huge attention from both academia and industry \cite{zhang2020empowering}. Generally, CVR modeling methods employ similar techniques developed for Click-Through Rate (CTR) prediction, which use high-order interactions of features to improve their representation capacity \cite{guo2017deepfm, wang2017deep, yu2020deep} and leverage sequential user behaviors to model users in a dynamic manner \cite{zhou2018deep, zhou2019deep, xiao2020deep}. However, due to label collection and dataset size problems, CVR modeling becomes quite different and challenging.

The major difficulties of CVR modeling are introduced by the well-known Sample Selection Bias (SSB) \cite{zadrozny2004learning} and Data Sparsity (DS) \cite{lee2012estimating} issues. Several studies have been carried out to tackle these issues following the idea of entire space modeling, $e.g.$, ESMM \cite{ma2018entire}, ESM$^2$ \cite{wen2020entire} and HM$^3$ \cite{wen2021hierarchically}. Besides, there exists a severe delayed feedback issue in CVR modeling, resulting in false negative labels in the CVR dataset. For this issue, ESDF \cite{wang2020delayed} combines the time delay factor and the advantage of entire space modeling, trying to solve the above three challenges at the same time.

However, most of the related works assume that data distribution of CVR samples is identical over time. For large-scale platforms such as Taobao and Amazon, this assumption is guaranteed by collecting training samples within a short time window, $e.g.$, a week or two, while in small scenarios usually several months of data are needed to train deep CVR models. The large time span may result in discrepant data distributions between training and online serving, which hurts the generalization performance. Besides, in e-commerce, users' shopping decisions can be influenced by different $occasions$ \cite{wang2020time}. Especially, promotions are being more prevalent and important to attract customers and boost sales, which have remarkable impacts on the data distribution within a short time window. In such situations, CVR modeling is highly challenging since the distribution of the upcoming occasion is uncertain and training samples of each occasion become very scarce. The distribution discrepancy and uncertainty problems mentioned above are summarized as \textbf{D}ata \textbf{D}istribution \textbf{F}luctuation (DDF) issue, which widely exists in small-scale recommendation scenarios while remaining under-explored in CVR prediction.

In this paper, we aim to tackle the DDF issue, which we consider as the primary problem in CVR modeling for small-scale scenarios, where most of the existing CVR methods can not perform well. After a detailed analysis of the logs, we observe that purchases on different occasions show different patterns. For example, people tend to purchase items of intrinsic preferences on normal days while they tend to engage with emerging hot items during promotions. Moreover, user behaviors before, during and after promotions also vary a lot. Inspired by this, we consider that samples of each class on each occasion cluster around a single prototype \cite{snell2017prototypical}. So we decompose the small scenario into four occasions, $i.e.$, Before-Promotion (BP), During-Promotion (DP), After-Promotion (AP) and Not-Promotion (NP). If all prototypes could be obtained and their distances to a query sample could be measured, we can tackle the distribution uncertainty accordingly. However, training samples on each occasion become further scarce after the decomposition. To achieve effective transfer learning across occasions with extremely limited data, we propose a novel CVR method named MetaCVR from a perspective of metric-based meta learning \cite{vinyals2016matching, snell2017prototypical,zhang2019category,hospedales2020meta}.

Concretely, we design a base CVR model and train it sufficiently with samples across months despite the non-identical distribution. Discarding the output layers, the base CVR model is used as Feature Representation Network (FRN) which generates positive and negative prototypes for each occasion with corresponding samples. Then we elaborately devise a Distance Metric Network (DMN) to compute the distances between each sample and all prototype representations so that a sample's priori conversion tendencies on different occasions are provided. At last, we develop an Ensemble Prediction Network (EPN) to incorporate the output of FRN and DMN and predict the final CVR score. With the FRN frozen, the DMN and EPN are trained with samples from the recent time period, therefore easing the distribution discrepancy introduced by samples a long time ago.

Our main contributions are summarized as follows:
\begin{itemize}
    \item To the best of our knowledge, we are the first to pay attention to the DDF issue in CVR modeling for small-scale recommendation scenarios. We introduce the idea of decomposing the small scenario into different occasions according to their distribution difference.
    \item We propose a novel method from a perspective of meta learning to tackle the challenges introduced by the DDF issue, which models each occasion with prototype representations. Taking advantage of meta learning, our model achieves effective knowledge sharing across occasions.
    \item Experiments on real-world datasets and online A/B test demonstrate the superiority of our model over representative methods. The code of MetaCVR has been made publicly available\footnote{https://github.com/AaronPanXiaoFeng/MetaCVR}.
\end{itemize}

\section{The Proposed Method}

\subsection{Motivation}
\begin{figure*}[htbp]
    \centering
    \includegraphics[width=1\linewidth]{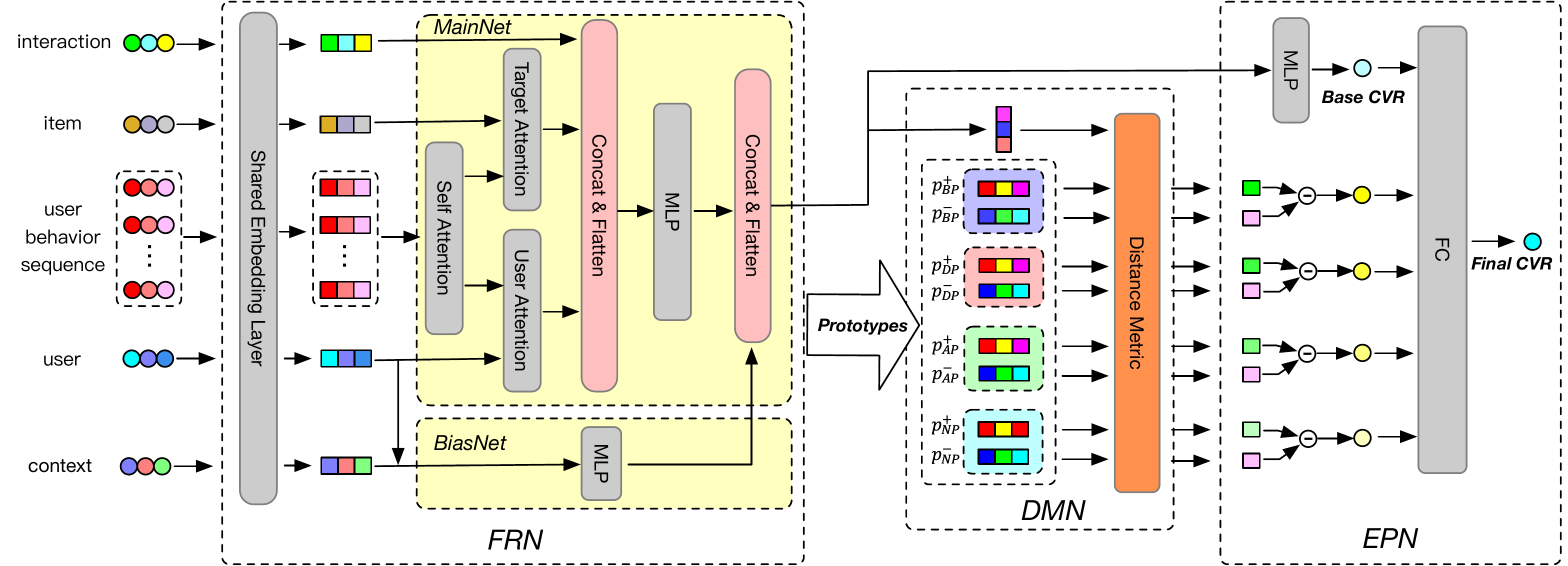}
    \caption{Framework of the proposed MetaCVR method. It contains a Feature Representation Network (FRN), a Distance Metric Network (DMN), and an Ensemble Prediction Network (EPN).}
    \label{fig:MetaL-CVR}
\end{figure*}

As illustrated in the introduction, we assume samples of each class on each occasion form a single prototype representation, $i.e.$, purchases driven by different occasions cluster around different patterns. Since different patterns coexist on each occasion with different impacts, simply distinguishing the source occasion of samples would not deliver good performance. Instead, we formulate the CVR prediction model parameterized by $\theta$ for small scenarios as:
\begin{equation} \label{eq:model}
    f_{\theta}(\bm{x})=g \left( \mathcal{F}(\bm{x}), \, \left\{ d \left( \mathcal{F}(\bm{x}), \bm{p}_{occ}^{cls} \right) \right\} \right),
\end{equation}
where $\mathcal{F}(\bm{x})$ denotes representation of input features $\bm{x}$ and $d(\cdot)$ denotes the distance metric function. With $cls \in \{+,-\}$ and $occ \in \{BP, DP, AP, NP\}$, $\bm{p}_{occ}^{cls}$ denotes the prototype representation of class $cls$ on occasion $occ$, and $g(\cdot)$ is the final prediction function. Symbol $+$ denotes the positive class while $-$ denotes the negative. In this paper, we implement $\mathcal{F}(\cdot)$, $d(\cdot)$ and $g(\cdot)$ as FRN, DMN and EPN respectively, which are shown in Figure~\ref{fig:MetaL-CVR} and will be detailed in the following sections.

\subsection{Base CVR Model}
\label{subsec:base}
As shown in Figure~\ref{fig:MetaL-CVR}, the input features of MetaCVR consist of five parts: 1) user behavior sequence; 2) user features including user profile and user statistic features; 3) item features such as item id, category, brand and related statistic features; 4) interaction features of the target item and user; 5) context features such as position and time information. All features are processed by a Shared Embedding Layer so we obtain the embedded user features, item features, interaction features, context features and user behavior sequence, $i.e.$, $\bm{e}^u$, $\bm{e}^i$, $\bm{e}^{ui}$, $\bm{e}^c$ and $\bm{e}^{seq}=\{\bm{e}^i_1,...,\bm{e}^i_t\}$, where $\bm{e}^i_t$ denotes the item embedding of $t^{th}$ user behavior and $t$ is the sequence length.

Then we perform three kinds of attention mechanisms in the MainNet of FRN. A multi-head self-attention \cite{vaswani2017attention,dosovitskiy2020image,xu2021vitae,zhang2022vitaev2} is calculated over $\bm{e}^{ubs}$ to model user preference from multiple views of interest and $\bm{\hat{e}}^{ubs}=\{\bm{\hat{e}}^i_1,...,\bm{\hat{e}}^i_t\}$ is the output. On top of the self-attention, user attention $\bm{s}^u$ is calculated to mine personalized information with $\bm{e}^{u}$ attending to $\bm{\hat{e}}^{ubs}$, and target attention $\bm{s}^i$ is employed to activate historical interests related to the target item with $\bm{e}^{i}$ attending to $\bm{\hat{e}}^{ubs}$. Then, $\bm{e}^{u}$, $\bm{e}^{i}$, $\bm{e}^{ui}$, $\bm{s}^i$ and $\bm{s}^u$ are concatenated and fed into a Multi-Layer Perception (MLP), generating $\bm{h}^M$ as the output of MainNet. To model the bias that different users in different contexts usually behave differently even to similar items, we feed $\bm{e}^{u}$ and $\bm{e}^{c}$ into another MLP (BiasNet), and $\bm{h}^B$ is the output.

Finally, we obtain the output of FRN and feed it to the output layers $f_b(\cdot)$, a ranking function implemented as a 3-layer MLP, to predict base CVR scores. The widely-used logloss is adopted as loss function to train the base CVR model, $i.e.$, 
\begin{equation} \label{eq:b_loss}
    \begin{aligned}
        \hat{y}_b&=f_b\left(\mathcal{F}(\bm{x})\right)=f_b\left(Norm([\bm{h}^M; \bm{h}^B])\right), \\
        l_b&=-\frac{1}{|\mathcal{D}|} \sum_{(\bm{x},y) \in \mathcal{D}} \left(y\,log\hat{y}_b + (1-y)\,log(1-\hat{y}_b)\right),
    \end{aligned}
\end{equation}
where $[;]$ refers to concatenation of vectors and $y \in \{0,1\}$ is the conversion label. In our small scenario, the training set $\mathcal{D}$ is collected from months of logs in order to train CVR models sufficiently. In contrast, $\mathcal{D}_r$ is constructed by logs of recent time period.

\subsection{Prototype Representations}
\label{subsec:proto}
Considering the differences of user behaviors on different occasions, we build support set for each occasion via picking a day of this occasion in $\mathcal{D}_r$ and splitting its samples into 2 subsets, $i.e.$, positive support set and negative support set. The positive support set includes all purchase samples of the day while the negative includes the rest of clicked samples. Then we employ the pre-trained FRN to map the input into a representation space and calculate the class's prototype as the mean feature of its support set, $i.e.$,
\begin{equation} \label{eq:proto}
    \begin{aligned}
        % V_{occ}^{cls}&=\left\{ \mathcal{F}(\bm{x}_k)\,|\,\bm{x}_k \in S_{occ}^{cls} \right\}, \\
        \bm{p}_{occ}^{cls}=Norm\left(Mean(V_{occ}^{cls})\right), \,\,V_{occ}^{cls}=\left\{ \mathcal{F}(\bm{x}_k)\,|\,\bm{x}_k \in S_{occ}^{cls} \right\},
    \end{aligned}
\end{equation}
where $S_{occ}^{cls}$ denotes the support set of class $cls$ on occasion $occ$, and $\bm{x}_k$ denotes input features of $k$th sample in $S_{occ}^{cls}$. In this way, we obtain 4 pairs of prototypes, $i.e.$, $\left\{ \{\bm{p}_{BP}^+,\bm{p}_{BP}^-\},\,\{\bm{p}_{DP}^+,\bm{p}_{DP}^-\},\,\{\bm{p}_{AP}^+,\bm{p}_{AP}^-\},\,\{\bm{p}_{NP}^+,\bm{p}_{NP}^-\} \right\}$.

\subsection{Distance Metric Network}
In metric-based meta learning, the choice of distance metric is crucial. In this paper, since the representation space of FRN is highly non-linear, it may not be suitable to choose fixed linear distance metrics such as cosine distance and Euclidean distance adopted in \cite{vinyals2016matching, snell2017prototypical}. We consider that a learnable distance metric can be a more generalizable solution and propose a trainable Space Projection Distance Metric (SPDM) which is formulated as follows:
\begin{equation} \label{eq:SPDM}
    \begin{aligned}
        d\left(\mathcal{F}(\bm{x}),\bm{p}_{occ}^{cls}\right)&=\mathcal{F}(\bm{x})^\mathrm{T} \bm{W}_{occ} \bm{p}_{occ}^{cls} + b_{occ},
    \end{aligned}
\end{equation}
where $\bm{W}_{occ}$ is a trainable projection matrix and $b_{occ}$ is a trainable bias scalar. It's worth mentioning that cosine distance is a special case of SPDM when $\bm{W}_{occ}$ is an identity matrix.

Inspired by Relation Network \cite{sung2018learning}, we also propose a Neural Network based Distance Metric (NNDM) which aims to learn the relation between query sample and support sets as a transferrable deep metric, $i.e.$,
\begin{equation} \label{eq:NNDM}
    \begin{aligned}
    d\left(\mathcal{F}(\bm{x}),\bm{p}_{occ}^{cls}\right)&=\bm{W}_{occ} \left[ \mathcal{F}(\bm{x}); \bm{p}_{occ}^{cls} \right] + b_{occ},
    \end{aligned}
\end{equation}

\subsection{Ensemble Prediction Network}
From SPDM or NNDM, we can obtain four pairs of distance metrics, $i.e.$, $\left\{ \{d_{BP}^+,d_{BP}^-\},\,\{d_{DP}^+,d_{DP}^-\},\,\{d_{AP}^+,d_{AP}^-\},\,\{d_{NP}^+,d_{NP}^-\} \right\}$. In most of the existing metric-based works, classification of a query sample is then performed by simply finding its nearest prototype, which is not directly applicable for CVR prediction since prototypes do not maintain fine-grained personalized information after mean pooling, which however is essential for a well-performing CVR model. Alternatively, we incorporate these distance metrics with the output of FRN in an ensemble approach, $i.e.$,
\begin{equation} \label{eq:ensemble}
    \begin{aligned}
        \hat{y}=f_e\left(s_b,\{s_{occ}\}\right),
        \,s_b=f_b(\mathcal{F}(\bm{x})), \,s_{occ}=d_{occ}^+-d_{occ}^-,
    \end{aligned}
\end{equation}
where $s_b$ denotes CVR prediction of the base CVR model and $s_{occ}$ represents how likely the purchase would happen on occasion $occ$. The final CVR score $\hat{y}$ is predicted by a fully connected layer, $i.e.$, $f_e(\cdot)$. Similarly to Eq.~\eqref{eq:b_loss}, logloss is adopted to train the whole model. Note that in this stage, we only train the parameters of DMN and EPN on $\mathcal{D}_r$ by stopping the gradient propagation to FRN.

\section{Experiments}

\subsection{Experimental Setup}
\subsubsection{\textbf{Datasets}}
By collecting logs\footnote{To the extent of our knowledge, there are no public datasets suited for the CVR task in settings of this paper.} between 2021/01/15 to 2021/03/31 from a small-scale recommendation scenario of our e-commerce platform, we establish the whole dataset which contains 6 promotions. The entire dataset is split into non-overlapped training set $\mathcal{D}$ (01/15-03/15) and validation set $\mathcal{D}_v$ (03/16-03/31). Besides, we build $\mathcal{D}_r$ with data from 03/01 to 03/15. Each of $\mathcal{D}$, $\mathcal{D}_v$ and $\mathcal{D}_r$ covers all occasions mentioned in this paper. Table~\ref{tab:dataset} summarizes their statistics. All the data have been anonymously processed and users' information is protected.

\begin{table}[t] %[htbp]
\centering
    \caption{Statistics of the established datasets.}
    \vspace{-3mm}
    \setlength{\tabcolsep}{1.5mm}{
        \begin{tabular}{c c c c c c}
            \hline
            \#Dataset & \#Users & \#Items & \#Exposures & \#Clicks & \#Purchases \\
            \hline
            $\mathcal{D}$ & 4.68M & 0.71M & 461.48M & 42.32M & 413K \\
            $\mathcal{D}_r$  & 2.62M & 0.68M & 121.37M & 12.38M & 141K \\
            $\mathcal{D}_v$ & 2.12M & 0.68M & 119.21M & 11.17M & 113K \\
            \hline
        \end{tabular}
    }
    \label{tab:dataset}
\end{table}

\subsubsection{\textbf{Evaluation Metrics}}
Area under ROC curve (AUC) is used as the offline evaluation metric. For online A/B testing, we choose PCVR$=p(conversion|click,impression)$ and GMV (Gross Merchandise Volume), which are widely adopted in industrial recommender systems. Improving PCVR and GMV simultaneously implies more accurate recommendations and business revenue growth.

\subsubsection{\textbf{Competitors}}
1) \textbf{XGBoost} \cite{chen2016xgboost} is a tree-based model which can produce competitive results for CVR prediction especially when samples are not enough for deep models. 2) \textbf{DCN} \cite{wang2017deep} applies feature crossing at each layer to capture high-order feature interactions to achieve better CVR prediction than DNN. 3) \textbf{BASE} refers to the base CVR model proposed in Section~\ref{subsec:base}, and \textbf{BASE-F} is the same as BASE except that it adopts the idea of fine-tuning from transfer learning \cite{caruana1995learning, bengio2012deep, yosinski2014transferable} to relieve the distribution discrepancy caused by the large time span. 4) \textbf{ESMM} \cite{ma2018entire} mitigates the SSB and DS issues by modeling CVR over the entire space. 5) \textbf{ESM$^2$} \cite{wen2020entire} extends ESMM and models purchase-related post-click behaviors in a unified multi-task learning framework. 6) \textbf{ESDF} \cite{wang2020delayed} combines the time delay factor and entire space modeling.

Features for all competitors are the same except that XGBoost and DCN discard user behavior sequence. For the XGBoost model, the number and depth of trees, minimum instance numbers for splitting a node, sampling rate of the training set and sampling rate of features for each iteration are set to 70, 5, 10, 0.6 and 0.6, respectively. All the deep models are implemented in distributed Tensorflow 1.4 and trained with 2 parameter severs and 3 Nvidia Tesla V100 16GB GPUs. Item ID, category ID and brand ID have an embedding size of 32 while 8 for the other categorical features. We use 8-head attention structures with a hidden size of 128. Adagrad optimizer with a learning rate of 0.01 and a mini-batch size of 256 are used for training. We report the results of each method under its empirically optimal hyper-parameters settings.

\begin{table}[t] %[htbp]
\centering
    \caption{Offline and online comparison results.}
    \vspace{-3mm}
    \setlength{\tabcolsep}{2.5mm}{
        \begin{tabular}{l c c c}
        \hline
        \multirow{2}{*}{Model} & Offline & \multicolumn{2}{c}{Online} \\
        % \cline{2-4}
        \cmidrule(r){2-2} \cmidrule(r){3-4}
        & AUC (mean$\pm$std.) & PCVR Gain & GMV Gain \\
        \hline
        XGBoost & 0.8264$\pm$0.00067 & 0\% & 0\% \\
        DCN & 0.8174$\pm$0.00074 & - & - \\
        BASE & 0.8233$\pm$0.00257 & -5.22\% & -6.58\% \\
        BASE-F & 0.8259$\pm$0.00187 & -1.01\% & -1.42\% \\
        ESMM & 0.8262$\pm$0.00163 & -0.83\% & -0.61\% \\
        ESM$^2$ & 0.8278$\pm$0.00151 & +2.36\% & +3.02\% \\
        ESDF & \underline{0.8291$\pm$0.00132} & \underline{+4.23\%} & \underline{+4.58\%} \\
        \textbf{MetaCVR} & \textbf{0.8395$\pm$0.00156} & \textbf{+11.92\%} & \textbf{+8.64\%} \\
        \hline
        \end{tabular}
    }
    \label{tab:comparison}
\end{table}

\subsection{Experimental Results}
\label{subsec:exps}
For offline evaluation, all experiments are repeated 3 times on $\mathcal{D}_v$. For online A/B testing, XGBoost was used as the baseline and the other models were tested in turn since the online traffic was not enough for testing all models simultaneously. Time for A/B testing of each model lasted more than 14 days, covering all occasions mentioned in this paper. The experimental results are presented in Table~\ref{tab:comparison}. The major observations are summarized as follows. \textbf{1)} XGBoost is a strong baseline for small scenarios, which outperforms all the deep models except for ESM$^2$, ESDF and MetaCVR. It requires fewer samples and thus alleviates the distribution discrepancy caused by the large time span. \textbf{2)} The BASE model outperforms the DCN model, validating its effectiveness. Moreover, BASE-F gains further improvements by fine-tuning the BASE model using recent samples, implying impacts of the distribution discrepancy do matter. \textbf{3)} ESMM is comparable to XGBoost and ESM$^2$ outperforms XGBoost by adopting more purchase-related signals. With the abundant supervisory signals introduced, the SSB and DS issues are mitigated and models are trained more sufficiently. \textbf{4)} ESDF further tackles the delayed feedback issue by providing more label correctness, becoming the runner-up method. \textbf{5)} Our MetaCVR model yields the best performance both offline and online, and outperforms the runner-up method by a large margin, confirming the impacts of the DDF issue in small scenarios and the effectiveness of our design.

Further, we explore the impacts of distance metrics by training 4 MetaCVR models using cosine distance, Euclidean distance, NNDM and SPDM in the DMN module respectively. The evaluation AUCs on $\mathcal{D}_v$ are \textbf{0.8294}, \textbf{0.8302}, \textbf{0.8346} and \textbf{0.8395} respectively. We can observe that performance achieved by different distance metric methods varies significantly, which confirms that the choice of distance metric is crucial. SPDM and NNDM outperform cosine and Euclidean obviously, validating that learnable distance metrics are better solutions especially when the representation space is unknown. Our proposed SPDM outperforms NNDM and achieves the best performance since the inductive bias of SPDM is simpler than NNDM when measuring distance metrics, which is beneficial in the context of limited data.

\subsection{Effectiveness analysis}
\begin{figure}[t]
\centering
    \subfigure[\scriptsize Negative prototypes]{
		\label{fig:proto_cosine_b}
		\includegraphics[width=0.48\linewidth]{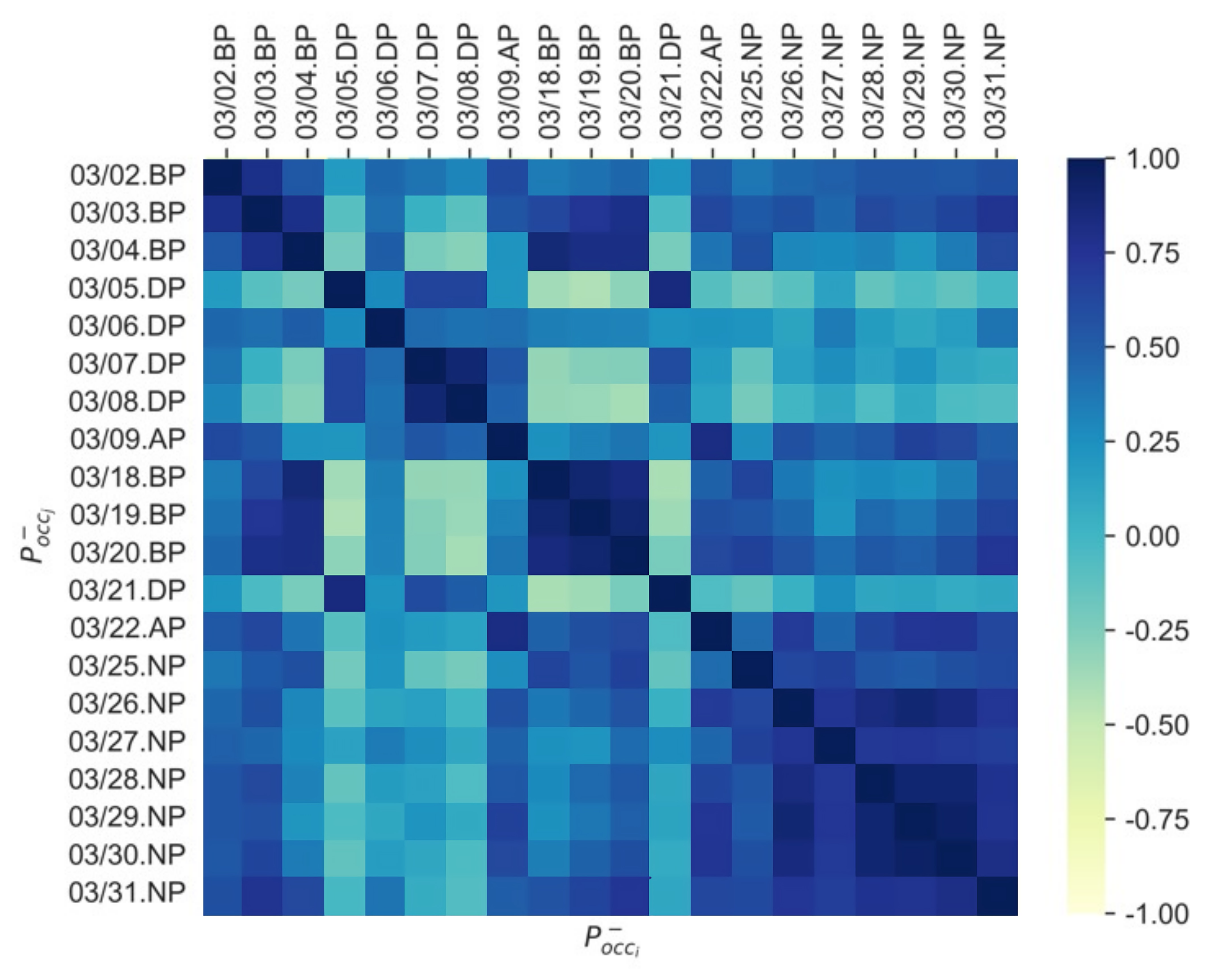}}
	%\hspace{0.1\linewidth} % gap
    \subfigure[\scriptsize Positive prototypes]{
		\label{fig:proto_cosine_a}
		\includegraphics[width=0.48\linewidth]{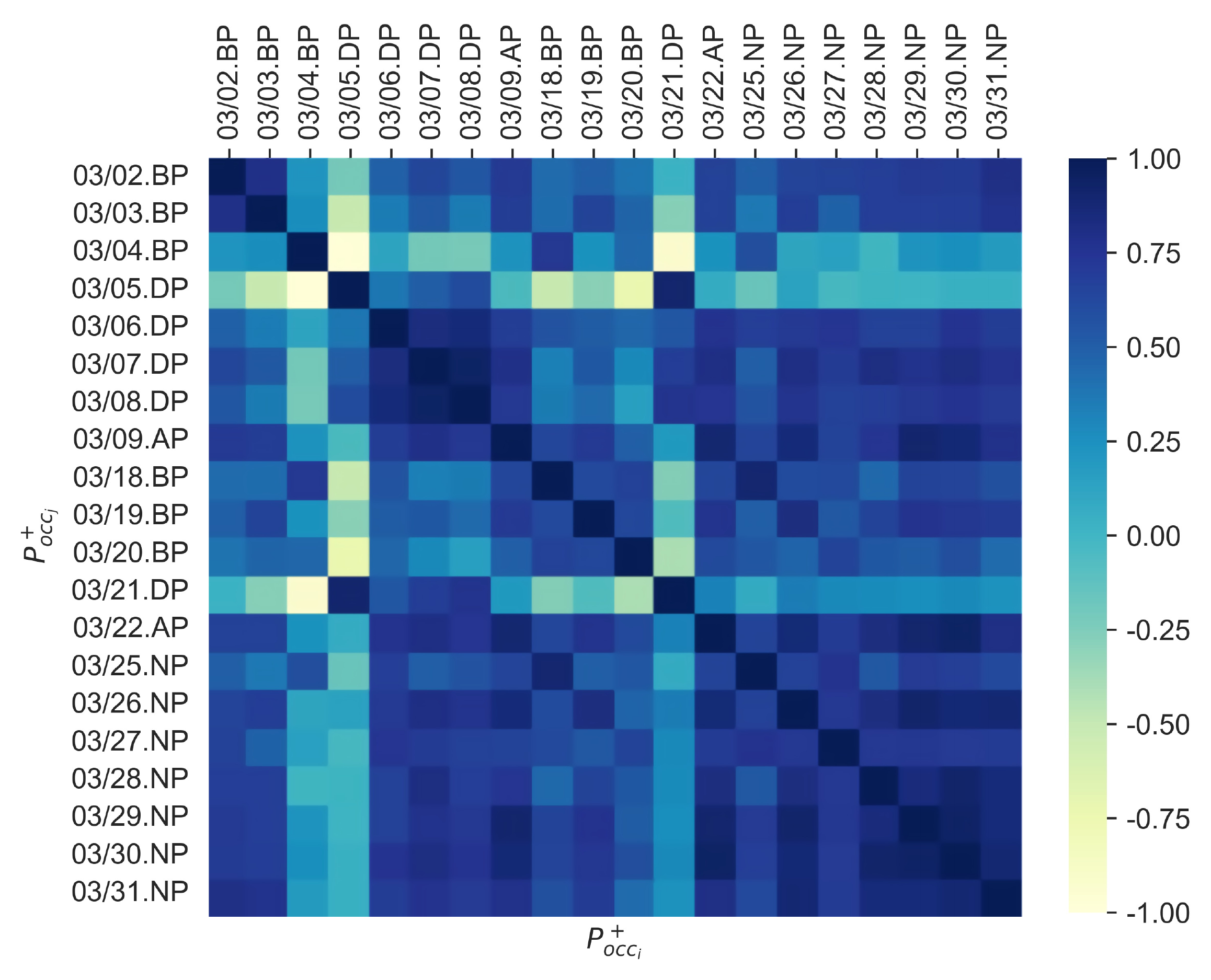}}
    \vspace{-3mm}
    \caption{Cosine similarity of prototype representations in different occasions.}
    \label{fig:proto_cosine}
\end{figure}

As summarized in Section~\ref{subsec:exps}, BASE-F performs better than BASE, and more impressive improvements are achieved when MetaCVR takes advantage of prototypes. Intuitively, BASE-F tackles the distribution discrepancy caused by the large time span, while MetaCVR partially employs this idea and further tackles the distribution uncertainty introduced by promotions, which we consider as the main reason for performance gains. Taking a step further, we investigate the effectiveness of prototypes. First, days of different occasions are picked from both $\mathcal{D}$ and $\mathcal{D}_v$, and prototypes of each chosen day are obtained according to Section~\ref{subsec:proto}. Then we compute cosine similarity of positive and negative prototypes respectively. As shown in Figure~\ref{fig:proto_cosine}, negative prototypes of the same occasion have high similarities while negative prototypes of different occasions have low similarities (positive prototypes behave similarly). The visualization empirically confirms our idea that prototypes of different occasions can distinguish from each other. With the learnable DMN better measuring distance, prototypes can be used to provide a sample's priori conversion tendencies on different occasions and thus help to handle the distribution uncertainty.

\section{Conclusion}
In this paper, we investigate the severe DDF issue in small-scale recommendation scenarios and propose a novel CVR prediction method named MetaCVR. It leverages the idea of metric-based meta learning to cope with the distribution uncertainty caused by frequent promotions and delivers promising transfer learning performance across occasions. Experiments on both offline datasets and online A/B test show MetaCVR significantly outperforms representative models. In the future, we intend to combine our method with entire space modeling and time delay factor, which is motivated by experimental results in Table~\ref{tab:comparison}.

%\bibliographystyle{ACM-Reference-Format}
%\bibliography{main}

\begin{thebibliography}{28}

%%% the Web address.

\ifx \showCODEN    \undefined \def \showCODEN     #1{\unskip}     \fi
\ifx \showDOI      \undefined \def \showDOI       #1{#1}\fi
\ifx \showISBNx    \undefined \def \showISBNx     #1{\unskip}     \fi
\ifx \showISBNxiii \undefined \def \showISBNxiii  #1{\unskip}     \fi
\ifx \showISSN     \undefined \def \showISSN      #1{\unskip}     \fi
\ifx \showLCCN     \undefined \def \showLCCN      #1{\unskip}     \fi
\ifx \shownote     \undefined \def \shownote      #1{#1}          \fi
\ifx \showarticletitle \undefined \def \showarticletitle #1{#1}   \fi
\ifx \showURL      \undefined \def \showURL       {\relax}        \fi
% invisible to TeX
\providecommand\bibfield[2]{#2}
\providecommand\bibinfo[2]{#2}
\providecommand\natexlab[1]{#1}
\providecommand\showeprint[2][]{arXiv:#2}

\bibitem[Bengio(2012)]%
        {bengio2012deep}
\bibfield{author}{\bibinfo{person}{Yoshua Bengio}.}
  \bibinfo{year}{2012}\natexlab{}.
\newblock \showarticletitle{Deep learning of representations for unsupervised
  and transfer learning}. In \bibinfo{booktitle}{\emph{Proceedings of ICML
  workshop on unsupervised and transfer learning}}. JMLR Workshop and
  Conference Proceedings, \bibinfo{pages}{17--36}.
\newblock


\bibitem[Caruana(1995)]%
        {caruana1995learning}
\bibfield{author}{\bibinfo{person}{Rich Caruana}.}
  \bibinfo{year}{1995}\natexlab{}.
\newblock \showarticletitle{Learning many related tasks at the same time with
  backpropagation}. In \bibinfo{booktitle}{\emph{Advances in neural information
  processing systems}}. \bibinfo{pages}{657--664}.
\newblock


\bibitem[Chen and Guestrin(2016)]%
        {chen2016xgboost}
\bibfield{author}{\bibinfo{person}{Tianqi Chen} {and} \bibinfo{person}{Carlos
  Guestrin}.} \bibinfo{year}{2016}\natexlab{}.
\newblock \showarticletitle{Xgboost: A scalable tree boosting system}. In
  \bibinfo{booktitle}{\emph{Proceedings of the 22nd acm sigkdd international
  conference on knowledge discovery and data mining}}.
  \bibinfo{pages}{785--794}.
\newblock


\bibitem[Dosovitskiy et~al\mbox{.}(2020)]%
        {dosovitskiy2020image}
\bibfield{author}{\bibinfo{person}{Alexey Dosovitskiy}, \bibinfo{person}{Lucas
  Beyer}, \bibinfo{person}{Alexander Kolesnikov}, \bibinfo{person}{Dirk
  Weissenborn}, \bibinfo{person}{Xiaohua Zhai}, \bibinfo{person}{Thomas
  Unterthiner}, \bibinfo{person}{Mostafa Dehghani}, \bibinfo{person}{Matthias
  Minderer}, \bibinfo{person}{Georg Heigold}, \bibinfo{person}{Sylvain Gelly},
  {et~al\mbox{.}}} \bibinfo{year}{2020}\natexlab{}.
\newblock \showarticletitle{An Image is Worth 16x16 Words: Transformers for
  Image Recognition at Scale}. In \bibinfo{booktitle}{\emph{International
  Conference on Learning Representations}}.
\newblock


\bibitem[Guo et~al\mbox{.}(2017)]%
        {guo2017deepfm}
\bibfield{author}{\bibinfo{person}{Huifeng Guo}, \bibinfo{person}{Ruiming
  Tang}, \bibinfo{person}{Yunming Ye}, \bibinfo{person}{Zhenguo Li}, {and}
  \bibinfo{person}{Xiuqiang He}.} \bibinfo{year}{2017}\natexlab{}.
\newblock \showarticletitle{DeepFM: a factorization-machine based neural
  network for CTR prediction}.
\newblock \bibinfo{journal}{\emph{arXiv preprint arXiv:1703.04247}}
  (\bibinfo{year}{2017}).
\newblock


\bibitem[Hospedales et~al\mbox{.}(2020)]%
        {hospedales2020meta}
\bibfield{author}{\bibinfo{person}{Timothy Hospedales},
  \bibinfo{person}{Antreas Antoniou}, \bibinfo{person}{Paul Micaelli}, {and}
  \bibinfo{person}{Amos Storkey}.} \bibinfo{year}{2020}\natexlab{}.
\newblock \showarticletitle{Meta-learning in neural networks: A survey}.
\newblock \bibinfo{journal}{\emph{arXiv preprint arXiv:2004.05439}}
  (\bibinfo{year}{2020}).
\newblock


\bibitem[Lee et~al\mbox{.}(2012)]%
        {lee2012estimating}
\bibfield{author}{\bibinfo{person}{Kuang-chih Lee}, \bibinfo{person}{Burkay
  Orten}, \bibinfo{person}{Ali Dasdan}, {and} \bibinfo{person}{Wentong Li}.}
  \bibinfo{year}{2012}\natexlab{}.
\newblock \showarticletitle{Estimating conversion rate in display advertising
  from past erformance data}. In \bibinfo{booktitle}{\emph{Proceedings of the
  18th ACM SIGKDD international conference on Knowledge discovery and data
  mining}}. ACM, \bibinfo{pages}{768--776}.
\newblock


\bibitem[Ma et~al\mbox{.}(2018)]%
        {ma2018entire}
\bibfield{author}{\bibinfo{person}{Xiao Ma}, \bibinfo{person}{Liqin Zhao},
  \bibinfo{person}{Guan Huang}, \bibinfo{person}{Zhi Wang},
  \bibinfo{person}{Zelin Hu}, \bibinfo{person}{Xiaoqiang Zhu}, {and}
  \bibinfo{person}{Kun Gai}.} \bibinfo{year}{2018}\natexlab{}.
\newblock \showarticletitle{Entire space multi-task model: An effective
  approach for estimating post-click conversion rate}. In
  \bibinfo{booktitle}{\emph{The 41st International ACM SIGIR Conference on
  Research \& Development in Information Retrieval}}. ACM,
  \bibinfo{pages}{1137--1140}.
\newblock


\bibitem[Snell et~al\mbox{.}(2017)]%
        {snell2017prototypical}
\bibfield{author}{\bibinfo{person}{Jake Snell}, \bibinfo{person}{Kevin
  Swersky}, {and} \bibinfo{person}{Richard Zemel}.}
  \bibinfo{year}{2017}\natexlab{}.
\newblock \showarticletitle{Prototypical networks for few-shot learning}.
\newblock \bibinfo{journal}{\emph{Advances in neural information processing
  systems}}  \bibinfo{volume}{30} (\bibinfo{year}{2017}).
\newblock


\bibitem[Sung et~al\mbox{.}(2018)]%
        {sung2018learning}
\bibfield{author}{\bibinfo{person}{Flood Sung}, \bibinfo{person}{Yongxin Yang},
  \bibinfo{person}{Li Zhang}, \bibinfo{person}{Tao Xiang},
  \bibinfo{person}{Philip~HS Torr}, {and} \bibinfo{person}{Timothy~M
  Hospedales}.} \bibinfo{year}{2018}\natexlab{}.
\newblock \showarticletitle{Learning to compare: Relation network for few-shot
  learning}. In \bibinfo{booktitle}{\emph{Proceedings of the IEEE conference on
  computer vision and pattern recognition}}. \bibinfo{pages}{1199--1208}.
\newblock


\bibitem[Vaswani et~al\mbox{.}(2017)]%
        {vaswani2017attention}
\bibfield{author}{\bibinfo{person}{Ashish Vaswani}, \bibinfo{person}{Noam
  Shazeer}, \bibinfo{person}{Niki Parmar}, \bibinfo{person}{Jakob Uszkoreit},
  \bibinfo{person}{Llion Jones}, \bibinfo{person}{Aidan~N Gomez},
  \bibinfo{person}{{\L}ukasz Kaiser}, {and} \bibinfo{person}{Illia
  Polosukhin}.} \bibinfo{year}{2017}\natexlab{}.
\newblock \showarticletitle{Attention is all you need}. In
  \bibinfo{booktitle}{\emph{Advances in neural information processing
  systems}}. \bibinfo{pages}{5998--6008}.
\newblock


\bibitem[Vinyals et~al\mbox{.}(2016)]%
        {vinyals2016matching}
\bibfield{author}{\bibinfo{person}{Oriol Vinyals}, \bibinfo{person}{Charles
  Blundell}, \bibinfo{person}{Timothy Lillicrap}, \bibinfo{person}{Daan
  Wierstra}, {et~al\mbox{.}}} \bibinfo{year}{2016}\natexlab{}.
\newblock \showarticletitle{Matching networks for one shot learning}.
\newblock \bibinfo{journal}{\emph{Advances in neural information processing
  systems}}  \bibinfo{volume}{29} (\bibinfo{year}{2016}),
  \bibinfo{pages}{3630--3638}.
\newblock


\bibitem[Wang et~al\mbox{.}(2020a)]%
        {wang2020time}
\bibfield{author}{\bibinfo{person}{Jianling Wang}, \bibinfo{person}{Raphael
  Louca}, \bibinfo{person}{Diane Hu}, \bibinfo{person}{Caitlin Cellier},
  \bibinfo{person}{James Caverlee}, {and} \bibinfo{person}{Liangjie Hong}.}
  \bibinfo{year}{2020}\natexlab{a}.
\newblock \showarticletitle{Time to Shop for Valentine's Day: Shopping
  Occasions and Sequential Recommendation in E-commerce}. In
  \bibinfo{booktitle}{\emph{Proceedings of the 13th International Conference on
  Web Search and Data Mining}}. \bibinfo{pages}{645--653}.
\newblock


\bibitem[Wang et~al\mbox{.}(2017)]%
        {wang2017deep}
\bibfield{author}{\bibinfo{person}{Ruoxi Wang}, \bibinfo{person}{Bin Fu},
  \bibinfo{person}{Gang Fu}, {and} \bibinfo{person}{Mingliang Wang}.}
  \bibinfo{year}{2017}\natexlab{}.
\newblock \showarticletitle{Deep \& cross network for ad click predictions}.
\newblock In \bibinfo{booktitle}{\emph{Proceedings of the ADKDD'17}}.
  \bibinfo{pages}{1--7}.
\newblock


\bibitem[Wang et~al\mbox{.}(2020b)]%
        {wang2020delayed}
\bibfield{author}{\bibinfo{person}{Yanshi Wang}, \bibinfo{person}{Jie Zhang},
  \bibinfo{person}{Qing Da}, {and} \bibinfo{person}{Anxiang Zeng}.}
  \bibinfo{year}{2020}\natexlab{b}.
\newblock \showarticletitle{Delayed feedback modeling for the entire space
  conversion rate prediction}.
\newblock \bibinfo{journal}{\emph{arXiv preprint arXiv:2011.11826}}
  (\bibinfo{year}{2020}).
\newblock


\bibitem[Wen et~al\mbox{.}(2019)]%
        {wen2019multi}
\bibfield{author}{\bibinfo{person}{Hong Wen}, \bibinfo{person}{Jing Zhang},
  \bibinfo{person}{Quan Lin}, \bibinfo{person}{Keping Yang}, {and}
  \bibinfo{person}{Pipei Huang}.} \bibinfo{year}{2019}\natexlab{}.
\newblock \showarticletitle{Multi-Level Deep Cascade Trees for Conversion Rate
  Prediction in Recommendation System}. In
  \bibinfo{booktitle}{\emph{Proceedings of the AAAI Conference on Artificial
  Intelligence}}.
\newblock


\bibitem[Wen et~al\mbox{.}(2021)]%
        {wen2021hierarchically}
\bibfield{author}{\bibinfo{person}{Hong Wen}, \bibinfo{person}{Jing Zhang},
  \bibinfo{person}{Fuyu Lv}, \bibinfo{person}{Wentian Bao},
  \bibinfo{person}{Tianyi Wang}, {and} \bibinfo{person}{Zulong Chen}.}
  \bibinfo{year}{2021}\natexlab{}.
\newblock \showarticletitle{Hierarchically Modeling Micro and Macro Behaviors
  via Multi-Task Learning for Conversion Rate Prediction}. In
  \bibinfo{booktitle}{\emph{Proceedings of the 44th International ACM SIGIR
  Conference on Research and Development in Information Retrieval}}.
\newblock


\bibitem[Wen et~al\mbox{.}(2020)]%
        {wen2020entire}
\bibfield{author}{\bibinfo{person}{Hong Wen}, \bibinfo{person}{Jing Zhang},
  \bibinfo{person}{Yuan Wang}, \bibinfo{person}{Fuyu Lv},
  \bibinfo{person}{Wentian Bao}, \bibinfo{person}{Quan Lin}, {and}
  \bibinfo{person}{Keping Yang}.} \bibinfo{year}{2020}\natexlab{}.
\newblock \showarticletitle{Entire space multi-task modeling via post-click
  behavior decomposition for conversion rate prediction}. In
  \bibinfo{booktitle}{\emph{Proceedings of the 43rd International ACM SIGIR
  Conference on Research and Development in Information Retrieval}}.
  \bibinfo{pages}{2377--2386}.
\newblock


\bibitem[Xiao et~al\mbox{.}(2020)]%
        {xiao2020deep}
\bibfield{author}{\bibinfo{person}{Zhibo Xiao}, \bibinfo{person}{Luwei Yang},
  \bibinfo{person}{Wen Jiang}, \bibinfo{person}{Yi Wei}, \bibinfo{person}{Yi
  Hu}, {and} \bibinfo{person}{Hao Wang}.} \bibinfo{year}{2020}\natexlab{}.
\newblock \showarticletitle{Deep Multi-Interest Network for Click-through Rate
  Prediction}. In \bibinfo{booktitle}{\emph{Proceedings of the 29th ACM
  International Conference on Information \& Knowledge Management}}.
  \bibinfo{pages}{2265--2268}.
\newblock


\bibitem[Xu et~al\mbox{.}(2021)]%
        {xu2021vitae}
\bibfield{author}{\bibinfo{person}{Yufei Xu}, \bibinfo{person}{Qiming Zhang},
  \bibinfo{person}{Jing Zhang}, {and} \bibinfo{person}{Dacheng Tao}.}
  \bibinfo{year}{2021}\natexlab{}.
\newblock \showarticletitle{Vitae: Vision transformer advanced by exploring
  intrinsic inductive bias}.
\newblock \bibinfo{journal}{\emph{Advances in Neural Information Processing
  Systems}}  \bibinfo{volume}{34} (\bibinfo{year}{2021}).
\newblock


\bibitem[Yosinski et~al\mbox{.}(2014)]%
        {yosinski2014transferable}
\bibfield{author}{\bibinfo{person}{Jason Yosinski}, \bibinfo{person}{Jeff
  Clune}, \bibinfo{person}{Yoshua Bengio}, {and} \bibinfo{person}{Hod Lipson}.}
  \bibinfo{year}{2014}\natexlab{}.
\newblock \showarticletitle{How transferable are features in deep neural
  networks?}
\newblock \bibinfo{journal}{\emph{arXiv preprint arXiv:1411.1792}}
  (\bibinfo{year}{2014}).
\newblock


\bibitem[Yu et~al\mbox{.}(2020)]%
        {yu2020deep}
\bibfield{author}{\bibinfo{person}{Feng Yu}, \bibinfo{person}{Zhaocheng Liu},
  \bibinfo{person}{Qiang Liu}, \bibinfo{person}{Haoli Zhang},
  \bibinfo{person}{Shu Wu}, {and} \bibinfo{person}{Liang Wang}.}
  \bibinfo{year}{2020}\natexlab{}.
\newblock \showarticletitle{Deep Interaction Machine: A Simple but Effective
  Model for High-order Feature Interactions}. In
  \bibinfo{booktitle}{\emph{Proceedings of the 29th ACM International
  Conference on Information \& Knowledge Management}}.
  \bibinfo{pages}{2285--2288}.
\newblock


\bibitem[Zadrozny(2004)]%
        {zadrozny2004learning}
\bibfield{author}{\bibinfo{person}{Bianca Zadrozny}.}
  \bibinfo{year}{2004}\natexlab{}.
\newblock \showarticletitle{Learning and evaluating classifiers under sample
  selection bias}. In \bibinfo{booktitle}{\emph{Proceedings of the
  international conference on Machine learning}}.
\newblock


\bibitem[Zhang and Tao(2021)]%
        {zhang2020empowering}
\bibfield{author}{\bibinfo{person}{Jing Zhang} {and} \bibinfo{person}{Dacheng
  Tao}.} \bibinfo{year}{2021}\natexlab{}.
\newblock \showarticletitle{Empowering Things With Intelligence: A Survey of
  the Progress, Challenges, and Opportunities in Artificial Intelligence of
  Things}.
\newblock \bibinfo{journal}{\emph{IEEE Internet of Things Journal}}
  \bibinfo{volume}{8}, \bibinfo{number}{10} (\bibinfo{year}{2021}),
  \bibinfo{pages}{7789--7817}.
\newblock
\urldef\tempurl%
\url{https://doi.org/10.1109/JIOT.2020.3039359}
\showDOI{\tempurl}


\bibitem[Zhang et~al\mbox{.}(2022)]%
        {zhang2022vitaev2}
\bibfield{author}{\bibinfo{person}{Qiming Zhang}, \bibinfo{person}{Yufei Xu},
  \bibinfo{person}{Jing Zhang}, {and} \bibinfo{person}{Dacheng Tao}.}
  \bibinfo{year}{2022}\natexlab{}.
\newblock \showarticletitle{ViTAEv2: Vision Transformer Advanced by Exploring
  Inductive Bias for Image Recognition and Beyond}.
\newblock \bibinfo{journal}{\emph{arXiv preprint arXiv:2202.10108}}
  (\bibinfo{year}{2022}).
\newblock


\bibitem[Zhang et~al\mbox{.}(2019)]%
        {zhang2019category}
\bibfield{author}{\bibinfo{person}{Qiming Zhang}, \bibinfo{person}{Jing Zhang},
  \bibinfo{person}{Wei Liu}, {and} \bibinfo{person}{Dacheng Tao}.}
  \bibinfo{year}{2019}\natexlab{}.
\newblock \showarticletitle{Category anchor-guided unsupervised domain
  adaptation for semantic segmentation}.
\newblock \bibinfo{journal}{\emph{Advances in Neural Information Processing
  Systems}}  \bibinfo{volume}{32} (\bibinfo{year}{2019}).
\newblock


\bibitem[Zhou et~al\mbox{.}(2019)]%
        {zhou2019deep}
\bibfield{author}{\bibinfo{person}{Guorui Zhou}, \bibinfo{person}{Na Mou},
  \bibinfo{person}{Ying Fan}, \bibinfo{person}{Qi Pi}, \bibinfo{person}{Weijie
  Bian}, \bibinfo{person}{Chang Zhou}, \bibinfo{person}{Xiaoqiang Zhu}, {and}
  \bibinfo{person}{Kun Gai}.} \bibinfo{year}{2019}\natexlab{}.
\newblock \showarticletitle{Deep interest evolution network for click-through
  rate prediction}. In \bibinfo{booktitle}{\emph{Proceedings of the AAAI
  conference on artificial intelligence}}, Vol.~\bibinfo{volume}{33}.
  \bibinfo{pages}{5941--5948}.
\newblock


\bibitem[Zhou et~al\mbox{.}(2018)]%
        {zhou2018deep}
\bibfield{author}{\bibinfo{person}{Guorui Zhou}, \bibinfo{person}{Xiaoqiang
  Zhu}, \bibinfo{person}{Chenru Song}, \bibinfo{person}{Ying Fan},
  \bibinfo{person}{Han Zhu}, \bibinfo{person}{Xiao Ma},
  \bibinfo{person}{Yanghui Yan}, \bibinfo{person}{Junqi Jin},
  \bibinfo{person}{Han Li}, {and} \bibinfo{person}{Kun Gai}.}
  \bibinfo{year}{2018}\natexlab{}.
\newblock \showarticletitle{Deep interest network for click-through rate
  prediction}. In \bibinfo{booktitle}{\emph{Proceedings of the 24th ACM SIGKDD
  International Conference on Knowledge Discovery \& Data Mining}}.
  \bibinfo{pages}{1059--1068}.
\newblock


\end{thebibliography}

% \appendix






\end{sloppypar}
\end{document}